\documentclass[10pt,twocolumn,letterpaper]{article}

\usepackage[pagenumbers]{cvpr} % To produce an arXiv/preprint version
\usepackage{multirow}
\usepackage{array}

\definecolor{cvprblue}{rgb}{0.21,0.49,0.74}
\usepackage[breaklinks,colorlinks,allcolors=cvprblue]{hyperref}

\def\paperID{*****} % *** Enter the Paper ID here
\def\confName{CVPR}
\def\confYear{2026}

\title{FAF-CD: Frequency-Aware Fusion for Change Detection under Imperfect Multimodal Remote Sensing}

\author{
Yufan Wang\textsuperscript{1} \quad
Sokratis Makrogiannis\textsuperscript{2} \quad
Chandra Kambhamettu\textsuperscript{1}\\
\textsuperscript{1}University of South Florida \quad
\textsuperscript{2}Delaware State University
}

\begin{document}
\maketitle

\begin{abstract}
Remote sensing change detection for real-world monitoring often relies on imperfect heterogeneous observations, where pre- and post-event images may be asynchronous, cross-sensor, or affected by illumination, seasonal, and modality shifts. This setting is especially challenging for EO-SAR disaster mapping, where nuisance variation can resemble structural damage. We propose \textbf{FAF-CD}, a frequency-aware hybrid framework with a DINOv3-pretrained ConvNeXt encoder and a linear-complexity VMamba-based decoder. Its rectification-aware tri-branch fusion module combines deformable spatial alignment with Fourier and Haar-wavelet comparisons, using adaptive gating to aggregate complementary cues across scales. On BRIGHT validation, a matched heterogeneous EO-SAR adaptation improves clean and perturbed tc-mIoU/tc-mAP over NeXt2Former-CD. FAF-CD also generalizes to binary optical CD, achieving 0.924 cF1 on LEVIR-CD and 0.955 cF1 on WHU-CD, and obtains the best average perturbed cIoU/cF1 on both binary datasets among M-CD and NeXt2Former-CD under pseudo-change-aligned stress tests. It further reduces cost by $\sim$24 GFLOPs relative to NeXt2Former-CD while maintaining or improving accuracy. Code and pretrained models are available at \url{https://github.com/VimsLab/FAF-CD}.
\end{abstract}
    
\section{Introduction}
\label{sec:intro}

Remote sensing image change detection (CD) aims to identify surface alterations between bi-temporal image pairs and plays a crucial role in applications such as urban planning, disaster monitoring, and land-use analysis. In practical monitoring and disaster-response settings, observations are rarely collected through an ideal, same-sensor lens: pre- and post-event images can be asynchronous, differ in resolution or viewpoint, and come from sensors with different imaging physics. Disaster response is a representative case, where pre-event electro-optical (EO) imagery may need to be compared with post-event synthetic aperture radar (SAR) because clouds, acquisition gaps, or emergency collection constraints prevent same-modality observation. These imperfect and heterogeneous acquisition conditions often create pseudo-changes, making it difficult to distinguish genuine structural modifications from appearance or modality-induced variation.

Modern CD frameworks predominantly adopt Siamese architectures to extract deep representations from bi-temporal imagery. Early convolutional neural networks (CNNs) provided strong spatial modeling capacity but struggled to capture long-range dependencies. Vision Transformers (ViTs)~\cite{dosovitskiy_image_2021} and attention-based models~\cite{chen_remote_2022, bandara_transformer-based_2022} were subsequently introduced to model global context more effectively, albeit with increased computational complexity. More recently, State Space Models (SSMs), particularly Mamba-based architectures~\cite{gu_mamba_2024}, have demonstrated the ability to achieve global receptive fields with linear complexity, leading to competitive CD models such as M-CD~\cite{paranjape_mamba-based_2024} and ChangeMamba~\cite{chen_changemamba_2024}.

Meanwhile, recent studies indicate that CNN backbones can remain competitive when paired with large-scale pretraining. In particular, NeXt2Former-CD~\cite{wang_next2former-cd_2026} reported that a Siamese ConvNeXt encoder~\cite{liu_convnet_2022} pretrained with DINOv3~\cite{simeoni_dinov3_2025} provides strong spatial features for CD. However, most existing methods, regardless of backbone choice, primarily rely on spatial-domain temporal fusion. Even when enhanced with deformable attention~\cite{zhu_deformable_2021}, spatial-only fusion can mix structural edges with environmental noise. Successive downsampling operations can also weaken high-frequency boundary details needed for delineating building footprints and small structural changes.

These observations motivate a complementary perspective: explicitly modeling temporal interactions in the frequency domain. Global frequency transformations, such as the Fast Fourier Transform (FFT)~\cite{yang_fda_2020}, help separate large-scale illumination and atmospheric variations from structural discrepancies. In parallel, localized frequency analysis via the 2D Discrete Wavelet Transform (2D-DWT)~\cite{yao_wave-vit_2022} provides directional high-frequency subbands that are particularly sensitive to edge-level discontinuities. Integrating such complementary frequency cues into temporal fusion offers a principled mechanism for disentangling appearance shifts from genuine geometric changes.

In this paper, we propose \textbf{FAF-CD}, a frequency-aware bi-temporal change detection framework designed for robust comparison under asymmetric acquisition and pseudo-change conditions. We adopt a Siamese DINOv3-pretrained ConvNeXt encoder~\cite{simeoni_dinov3_2025, liu_convnet_2022} to extract robust multi-scale representations. To address the limitations of purely spatial fusion, we design a rectification-aware tri-branch fusion module that integrates (i) deformable spatial alignment, (ii) global Fourier-based discrepancy modeling, and (iii) Haar wavelet-based boundary enhancement. An adaptive gating mechanism dynamically aggregates these complementary representations. The fused feature pyramid is then decoded by a VMamba-based decoder~\cite{liu_vmamba_2024, paranjape_mamba-based_2024}, enabling efficient global reasoning at high resolution. Beyond binary optical CD, we instantiate matched heterogeneous EO-SAR variants on BRIGHT to evaluate the framework under pre/post modality differences.

Our main contributions are summarized as follows:
\begin{itemize}
\item We present \textbf{FAF-CD}, a frequency-aware hybrid framework that combines a DINOv3-initialized Siamese ConvNeXt encoder with a VMamba-based decoder for accurate and efficient change detection.
\item We introduce a rectification-aware tri-branch fusion module that jointly models temporal interactions in spatial, Fourier, and wavelet domains, and adaptively aggregates them via a learnable gate.
\item We demonstrate the effectiveness of FAF-CD on BRIGHT~\cite{chen2025bright}, LEVIR-CD~\cite{chen_spatial-temporal_2020}, and WHU-CD~\cite{ji_fully_2019}, covering heterogeneous EO-SAR disaster mapping, clean binary optical CD, pseudo-change-aligned perturbations, and modality-side stress tests.
\end{itemize}

\section{Related Work}
\label{sec:relatedwork}

\subsection{Deep Learning-Based Change Detection}
Remote sensing change detection has moved from hand-crafted pixel-level comparisons toward learned feature representations. Many recent methods use Siamese networks, where two weight-shared branches encode the bi-temporal images before temporal comparison. Fully convolutional networks, often based on U-Net-like designs, provide effective local spatial features but can miss long-range context and confuse off-nadir displacement or seasonal appearance shifts with real changes. Attention mechanisms and Vision Transformers (ViTs)~\cite{dosovitskiy_image_2021} were introduced to model broader spatial context. Representative models such as the Bitemporal Image Transformer (BIT)~\cite{chen_remote_2022} and ChangeFormer~\cite{bandara_transformer-based_2022} use token- or self-attention-based context modeling for CD, although standard self-attention becomes expensive as image resolution increases.

\subsection{Foundation Models in Remote Sensing}

State Space Models (SSMs), including Mamba-based architectures~\cite{gu_mamba_2024, liu_vmamba_2024}, provide an alternative to dense self-attention by modeling long sequences with linear complexity. In remote sensing, recent work has adapted SSMs to classification and change detection. RSMamba~\cite{chen_rsmamba_2024} applies this family to remote sensing classification, while ChangeMamba~\cite{chen_changemamba_2024}, CDMamba~\cite{zhang_cdmamba_2025}, and M-CD~\cite{paranjape_mamba-based_2024} study Mamba-style temporal modeling for binary CD. Later variants explore different design choices: ChessMamba~\cite{ding_chessmamba_2025} interleaves state-space sequences, 3D-SSM~\cite{huang_3d-ssm_2025} uses a 3D selective scan module, and FA-Mamba~\cite{sun_fa-mamba_2025} and CMSNet~\cite{zhang_cmsnet_2025} combine SSM blocks with feature augmentation or foundation-model priors. Other related work uses synthetic data augmentation (SAMBA)~\cite{tobias_samba_2024} or boundary-specific supervision~\cite{wang_change_2025}.

These SSM-based methods show the value of efficient long-range modeling, but recent convolutional backbones remain competitive when paired with large-scale pretraining. NeXt2Former-CD~\cite{wang_next2former-cd_2026} uses a Siamese ConvNeXt~\cite{liu_convnet_2022} encoder initialized with DINOv3~\cite{simeoni_dinov3_2025} and reports strong results against recent SSM-based CD models. Our framework follows this encoder choice and uses a VMamba-based~\cite{liu_vmamba_2024} decoder, related to the decoder design in M-CD~\cite{paranjape_mamba-based_2024}, to retain efficient global context modeling during mask prediction.

\begin{figure*}[t]
    \centering
    \includegraphics[width=\linewidth]{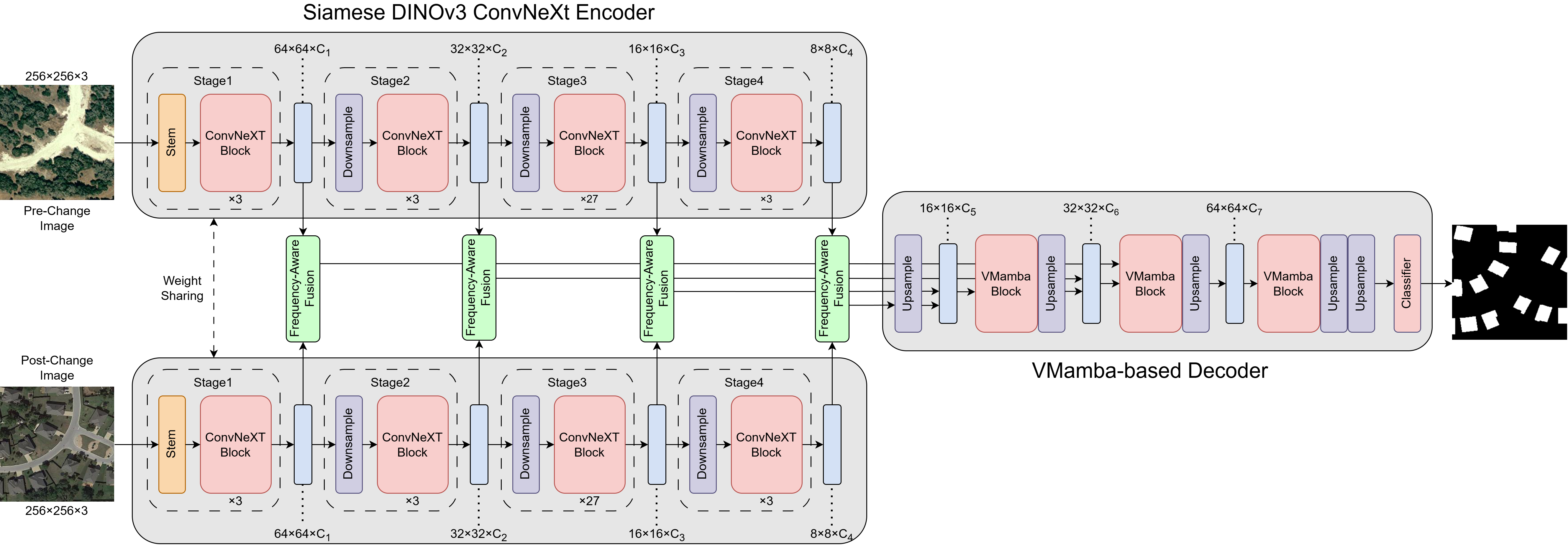}
    \caption{
        Overview of the proposed end-to-end change detection framework in the binary optical CD setting. Given a bi-temporal image pair, a weight-sharing Siamese ConvNeXt-L~\cite{liu_convnet_2022} encoder initialized with DINOv3~\cite{simeoni_dinov3_2025} pretraining extracts multi-scale features at four stages. At each stage, we apply a frequency-aware temporal fusion module to align and compare pre- and post-change representations, producing fused features that emphasize change-relevant structure. The fused feature pyramid is then fed into a VMamba-based~\cite{liu_vmamba_2024} decoder with progressive upsampling to generate dense change logits, followed by a lightweight classifier to output the final binary change mask. The BRIGHT adaptation uses pseudo-Siamese ConvNeXt-Base branches and a four-class building-status head.
    }
    \label{fig:architecture}
\end{figure*}

\subsection{Feature Fusion in Remote Sensing}

Feature fusion is a central component of Siamese CD models because the network must compare bi-temporal features while reducing nuisance variation. Common spatial fusion strategies concatenate features or use absolute differences, but these operations do not explicitly handle misalignment or boundary detail. More recent spatial fusion modules align features across time and scale. For example, the Feature Rectify Modules (FRM) and Feature Fusion Modules (FFM) in NeXt2Former-CD~\cite{wang_next2former-cd_2026} use deformable attention~\cite{zhu_deformable_2021} to sample spatial locations adaptively for bi-temporal alignment and cross-scale interaction. However, spatial-domain fusion alone can still struggle to separate structural changes from background appearance shifts.

Frequency-domain analysis provides another way to model appearance variation and boundary structure. In general vision, Fourier Domain Adaptation (FDA)~\cite{yang_fda_2020} uses frequency statistics to reduce domain discrepancies, while MPNet~\cite{wei_multiscale_2025} and FDFENet~\cite{he_fdfenet_2026} apply spatial-frequency fusion to remote sensing CD. Wavelet transforms provide localized frequency subbands that can represent edge and texture changes. Wave-ViT~\cite{yao_wave-vit_2022} and MLWNet~\cite{gao_efficient_2024} study wavelet-based operations in broader vision tasks, and recent CD models such as WGDF~\cite{zhang_wavelet-guided_2025} and BWFNet~\cite{lu_bwfnet_2025} use wavelet-frequency features for change representation. Building on these observations, our framework integrates Real Fast Fourier Transform (FFT) and Haar 2D Discrete Wavelet Transform (2D-DWT) into the temporal fusion pipeline to complement spatial alignment with global frequency and local edge cues.

\section{Method}
\label{sec:method}

We propose FAF-CD, an end-to-end bi-temporal change detection network that integrates a Siamese DINOv3 encoder, a rectification-aware multi-branch fusion mechanism, and a VMamba-based decoder. As illustrated in \cref{fig:architecture}, the framework follows the Siamese design used in recent CD models~\cite{paranjape_mamba-based_2024,wang_next2former-cd_2026} and adds explicit frequency-aware fusion to improve robustness to appearance shifts and texture variations.

\begin{figure}
    \centering
    \includegraphics[width=0.62\linewidth]{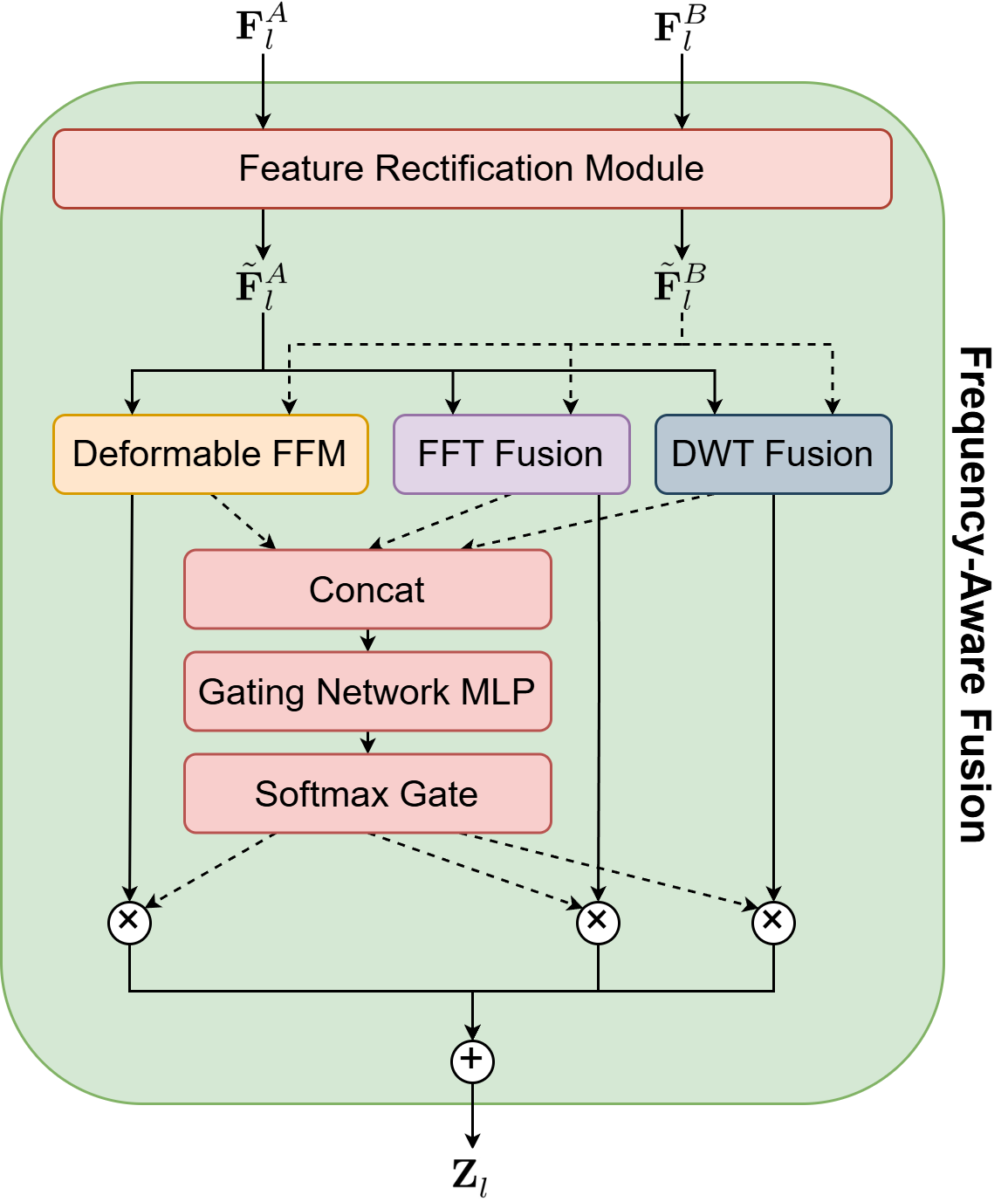}
    \caption{Detailed architecture of the Frequency-Aware Fusion module. Bitemporal features are first aligned via the Feature Rectification Module, then processed in parallel across spatial (Deformable FFM), global frequency (FFT), and local high-frequency (DWT) domains. An adaptive gating network aggregates these multi-domain representations into a unified feature map $\mathbf{Z}_l$.}
    \label{fig:freq_aware_fusion}
\end{figure}

\subsection{Problem Formulation}
For the standard binary optical CD setting, given a bi-temporal image pair
\begin{equation}
\mathbf{X}^{A}, \mathbf{X}^{B} \in \mathbb{R}^{3 \times H \times W},
\end{equation}
the goal is to predict a binary change map
\begin{equation}
\mathbf{Y} \in \{0,1\}^{H \times W}.
\end{equation}
Our network learns the mapping function
\begin{equation}
\mathbf{Y} = \mathcal{F}\!\left(\mathbf{X}^{A}, \mathbf{X}^{B}\right).
\end{equation}
The BRIGHT validation setting in \cref{sec:experiments} uses a dataset-specific four-class multimodal variant of the same framework.

\subsection{Siamese DINOv3 Encoder}
To obtain multi-scale visual features, we adopt a ConvNeXt encoder initialized with DINOv3 priors~\cite{simeoni_dinov3_2025, liu_convnet_2022}. The main LEVIR-CD and WHU-CD experiments use a weight-sharing Siamese ConvNeXt-L encoder, while the BRIGHT adaptation uses pseudo-Siamese ConvNeXt-Base branches initialized from the same DINOv3 checkpoint to handle different pre- and post-event modalities. Compared with models trained only for supervised classification, the DINOv3 foundation model provides dense, fine-grained visual descriptors that can improve transfer under domain shifts. For each stage $l \in \{1,2,3,4\}$, the encoder outputs a feature map:
\begin{equation}
\mathbf{F}^{A}_{l},\mathbf{F}^{B}_{l} \in \mathbb{R}^{C_l \times H_l \times W_l}.
\end{equation}
The shared or jointly initialized temporal branches map the two observations into a comparable feature space for downstream temporal comparison.

\subsection{Rectification and Tri-Branch Fusion}
A purely spatial comparison often entangles genuine structural changes with environmental noise. To decouple these factors, we process the features at each stage $l$ through a Rectification-Aware Tri-Branch Fusion module, as depicted in \cref{fig:freq_aware_fusion}.

\paragraph{Feature Rectification.}
We first mitigate spatial misalignment caused by varying sensor angles or perspective shifts. Following the Feature Rectification Module (FRM) design in NeXt2Former-CD~\cite{wang_next2former-cd_2026}, we spatially calibrate the raw encoder outputs:
\begin{equation}
\left(\tilde{\mathbf{F}}^{A}_{l},\tilde{\mathbf{F}}^{B}_{l}\right)
=
\mathrm{FRM}\!\left(\mathbf{F}^{A}_{l},\mathbf{F}^{B}_{l}\right).
\end{equation}

\paragraph{Branch 1: Deformable FFM.}
The first branch addresses standard spatial-domain alignment. Following the Feature Fusion Module (FFM) design in NeXt2Former-CD~\cite{wang_next2former-cd_2026}, this branch applies deformable cross-attention~\cite{zhu_deformable_2021} to adaptively sample relevant contextual locations:
\begin{equation}
\mathbf{Z}^{\mathrm{ffm}}_{l}
=
\mathrm{FFM}\!\left(\tilde{\mathbf{F}}^{A}_{l},\tilde{\mathbf{F}}^{B}_{l}\right).
\end{equation}

\paragraph{Branch 2: RFFT Fusion.}
To suppress global environmental pseudo-changes (e.g., broad illumination shifts), the second branch maps the temporal features into the frequency domain via a 2D Real Fast Fourier Transform (rFFT2):
\begin{equation}
\mathcal{S}^{A}_{l}=\mathrm{rFFT2}\!\left(\tilde{\mathbf{F}}^{A}_{l}\right),\quad
\mathcal{S}^{B}_{l}=\mathrm{rFFT2}\!\left(\tilde{\mathbf{F}}^{B}_{l}\right).
\end{equation}
We compute both the pixel-wise average and the absolute difference of the spectral representations:
\begin{equation}
\mathcal{S}^{\mathrm{mean}}_{l}=\frac{\mathcal{S}^{A}_{l}+\mathcal{S}^{B}_{l}}{2},\quad
\mathcal{S}^{\Delta}_{l}=|\mathcal{S}^{A}_{l}-\mathcal{S}^{B}_{l}|.
\end{equation}
The average spectrum preserves shared global context, while the difference spectrum highlights domain discrepancies. Following the inverse transform (irFFT2), the components are concatenated and refined through a lightweight block consisting of a $1\times1$ convolution, Batch Normalization, ReLU activation, and a $3\times3$ convolution to yield the final Fourier-fused feature $\mathbf{Z}^{\mathrm{fft}}_{l}$.

\paragraph{Branch 3: Haar 2D-DWT Fusion.}
The third branch injects localized high-frequency cues via a fixed Haar 2D-DWT.
We apply the transform channel-wise using grouped convolution with stride $2$ to obtain four subbands at half resolution:
\begin{equation}
(\mathbf{LL}^{t}_{l},\mathbf{LH}^{t}_{l},\mathbf{HL}^{t}_{l},\mathbf{HH}^{t}_{l})
=\mathrm{DWT}_{\mathrm{Haar}}(\tilde{\mathbf{F}}^{t}_{l}),\quad t\in\{A,B\}.
\end{equation}
For odd $H_l$ or $W_l$, we use replicate padding to ensure even spatial sizes.
We fuse subbands following our implementation: the low-frequency component is averaged,
\begin{equation}
\mathbf{L}_l=\tfrac{1}{2}\left(\mathbf{LL}^{A}_{l}+\mathbf{LL}^{B}_{l}\right),
\end{equation}
and the high-frequency response is aggregated by summing directional absolute differences,
\begin{equation}
\mathbf{H}_l=
|\mathbf{LH}^{A}_{l}-\mathbf{LH}^{B}_{l}|
+
|\mathbf{HL}^{A}_{l}-\mathbf{HL}^{B}_{l}|
+
|\mathbf{HH}^{A}_{l}-\mathbf{HH}^{B}_{l}|.
\end{equation}
We bilinearly upsample $\mathbf{L}_l$ and $\mathbf{H}_l$ back to $(H_l,W_l)$, concatenate, and apply a lightweight projection to obtain $\mathbf{Z}^{\mathrm{dwt}}_{l}$.

\paragraph{Adaptive Gated Aggregation.}
The outputs from the three branches represent distinct views of the bi-temporal change. As shown at the bottom of \cref{fig:freq_aware_fusion}, these features are concatenated and fed into an MLP-based gating network. A Softmax function yields adaptive branch-wise weights:
\begin{equation}
[\alpha_l,\beta_l,\gamma_l]
=
\mathrm{Softmax}\!\left(
\frac{\phi_l\!\left([\mathbf{Z}^{\mathrm{ffm}}_{l};\mathbf{Z}^{\mathrm{fft}}_{l};\mathbf{Z}^{\mathrm{dwt}}_{l}]\right)}{\tau}
\right),
\end{equation}
with $\alpha_l+\beta_l+\gamma_l=1$. The final fused feature for stage $l$ is obtained via a weighted summation:
\begin{equation}
\mathbf{Z}_{l}
=
\alpha_l\mathbf{Z}^{\mathrm{ffm}}_{l}
+
\beta_l\mathbf{Z}^{\mathrm{fft}}_{l}
+
\gamma_l\mathbf{Z}^{\mathrm{dwt}}_{l}.
\end{equation}

\subsection{VMamba-Based Decoder}
The resulting fused multi-scale pyramid $\{\mathbf{Z}_{l}\}_{l=1}^{4}$ is routed to a VMamba-based decoder. Leveraging the state-space decoding strategy introduced by recent architectures~\cite{paranjape_mamba-based_2024,liu_vmamba_2024}, the decoder progressively upsamples the representations while maintaining a global effective receptive field. This yields dense, high-resolution change logits:
\begin{equation}
\mathbf{P} \in \mathbb{R}^{K \times H \times W},
\end{equation}
upon which pixel-wise classification is applied to generate the final mask. Here $K=2$ for binary optical CD and $K=4$ for the BRIGHT building-status adaptation.

\subsection{Optimization Objective}
The network is optimized end-to-end using a weighted cross-entropy loss to address class imbalance:
\begin{equation}
\mathcal{L}_{\mathrm{ce}}
=
-\frac{1}{|\Omega|}
\sum_{u\in\Omega}\sum_{c=1}^{K}
w_c\,\mathbf{1}(y_u=c)\log p_{u,c},
\end{equation}
where $\Omega$ denotes the set of all spatial pixels, $w_c$ is the penalty weight for class $c$, and $p_{u,c}$ is the predicted probability for class $c$ at pixel location $u$.

\section{Experiments}
\label{sec:experiments}

\newcommand{\rr}{\raggedright\arraybackslash}
\begin{table*}
  \centering
  \renewcommand{\arraystretch}{1.15}
  \setlength{\tabcolsep}{3pt}
  \resizebox{\textwidth}{!}{%
  \begin{tabular}{>{\rr}p{4.2cm} c c c c c c c c c c}

    \toprule
    & & \multicolumn{3}{c}{WHU-CD~\cite{ji_fully_2019}} & \multicolumn{3}{c}{LEVIR-CD~\cite{chen_spatial-temporal_2020}} & \multicolumn{3}{c}{BRIGHT val~\cite{chen2025bright}} \\
    \cmidrule(lr){3-5}\cmidrule(lr){6-8}\cmidrule(lr){9-11}
    \multirow{-2}{*}{Method} & \multirow{-2}{*}{Extra training data}
      & cF1 ($\uparrow$) & cIoU ($\uparrow$) & OA ($\uparrow$)
      & cF1 ($\uparrow$) & cIoU ($\uparrow$) & OA ($\uparrow$)
      & tc-mIoU ($\uparrow$) & tc-mAP ($\uparrow$) & OA ($\uparrow$) \\
    \midrule

    \multicolumn{11}{l}{\textcolor{blue}{CNN-based Methods:}}\\
    SNUNet~\cite{fang_snunet-cd_2022} & None
      & 0.835 & 0.717 & 98.7
      & 0.882 & 0.788 & 98.8
      & -- & -- & -- \\
    IFNet~\cite{zhang_deeply_2020} & ImageNet 1k
      & 0.834 & 0.715 & 98.8
      & 0.881 & 0.788 & 98.9
      & -- & -- & -- \\
    \midrule

    \multicolumn{11}{l}{\textcolor{blue}{CNN + Attention based Methods:}}\\
    DT-SCN~\cite{liu_building_2021} & ImageNet 1k
      & 0.914 & 0.842 & 99.3
      & 0.877 & 0.781 & 98.8
      & -- & -- & -- \\
    \midrule

    \multicolumn{11}{l}{\textcolor{blue}{Transformer-based Methods:}}\\
    BIT~\cite{chen_remote_2022} & ImageNet 1k
      & 0.905 & 0.834 & 99.3
      & 0.893 & 0.807 & 98.92
      & -- & -- & -- \\
    ChangeFormer~\cite{bandara_transformer-based_2022} & None
      & 0.886 & 0.795 & 99.1
      & 0.904 & 0.825 & 99.0
      & -- & -- & -- \\
    \midrule

    \multicolumn{11}{l}{\textcolor{blue}{Diffusion-based Methods}}\\
    DDPM-CD~\cite{bandara_ddpm-cd_2024} & Google Earth
      & 0.927 & 0.863 & 99.4
      & 0.909 & 0.833 & 99.1
      & -- & -- & -- \\
    \midrule

    \multicolumn{11}{l}{\textcolor{blue}{Mamba-based Methods}}\\
    RSMamba~\cite{chen_rsmamba_2024} & ImageNet 1k
      & 0.927 & 0.865 & 99.4
      & 0.897 & 0.814 & 98.9
      & -- & -- & -- \\
    ChangeMamba~\cite{chen_changemamba_2024}& ImageNet 1k
      & 0.925 & 0.861 & 99.4
      & 0.902 & 0.821 & 99.0
      & -- & -- & -- \\
    CDMamba~\cite{zhang_cdmamba_2025} & ImageNet 1k
      & 0.937 & 0.882 & 99.5
      & 0.907 & 0.831 & 99.0
      & -- & -- & -- \\
    M-CD~\cite{paranjape_mamba-based_2024} & ImageNet 1k
      & \underline{0.954} & \underline{0.911} & 99.635
      & 0.919 & 0.850 & 99.185
      & -- & -- & -- \\
    \midrule

    \multicolumn{11}{l}{\textcolor{blue}{Hybrid Methods}}\\

    NeXt2Former-CD~\cite{wang_next2former-cd_2026} & LVD-1689M
      & \textbf{0.955} & \textbf{0.914} & \textbf{99.646}
      & \underline{0.921} & \underline{0.854} & \underline{99.202}
      & \underline{0.5474} & \underline{0.7479} & \underline{94.98} \\

    \midrule

    Ours & LVD-1689M
      & \textbf{0.955} & \textbf{0.914} & \underline{99.642}
      & \textbf{0.924} & \textbf{0.859} & \textbf{99.233}
      & \textbf{0.5846} & \textbf{0.7691} & \textbf{96.65} \\

    \bottomrule
  \end{tabular}}
    \caption{
        Comparison of our proposed model with state-of-the-art change detection methods.
        cF1 and cIoU denote the F1 score and Intersection-over-Union for the binary change class, and OA denotes overall accuracy.
        BRIGHT results are validation-only measurements in the heterogeneous EO-SAR disaster-mapping setting using BRIGHT-adapted pseudo-Siamese ConvNeXt-Base variants of NeXt2Former-CD and FAF-CD; tc-mIoU and tc-mAP denote target-class mean IoU and detection-style AP over intact, damaged, and destroyed, excluding background.
        As with cF1/cIoU, BRIGHT tc-mIoU/tc-mAP are reported as fractions, while OA is reported as a percentage.
        For LEVIR-CD and WHU-CD, we use the reported NeXt2Former-CD results; for BRIGHT, we train a matched NeXt2Former-CD adaptation for the validation-only EO-SAR task.
        For M-CD~\cite{paranjape_mamba-based_2024}, we are reporting the reevaluated cF1 scores following~\cite{wang_next2former-cd_2026}.
        The best result is indicated in \textbf{bold}, and the second-best result is \underline{underlined}.
    }

  \label{tab:mcd_sota_comparison}
\end{table*}

\subsection{Datasets}
We evaluate the proposed FAF-CD framework on two widely adopted remote sensing change detection benchmarks: LEVIR-CD~\cite{chen_spatial-temporal_2020} and WHU-CD~\cite{ji_fully_2019}. LEVIR-CD comprises high-resolution Google Earth imagery focusing on the construction and demolition of large-scale buildings, providing 7,120 bi-temporal patch pairs for training, 1,024 for validation, and 2,048 for testing. WHU-CD~\cite{ji_fully_2019} is a building change detection benchmark constructed from high-resolution aerial imagery over urban areas, and we use 5{,}947 training, 743 validation, and 744 testing pairs following the official split.

To ensure strict comparability with recent literature, we utilize the exact data preprocessing and official splits distributed by prior work~\cite{paranjape_mamba-based_2024}, where the original high-resolution bi-temporal images are cropped into non-overlapping $256 \times 256$ spatial patches. For these binary CD benchmarks, we report cF1 and cIoU for the change category, as well as the overall pixel accuracy (OA), computed exclusively on the test sets.

To evaluate imperfect multimodal monitoring, we additionally evaluate NeXt2Former-CD~\cite{wang_next2former-cd_2026} and FAF-CD on the BRIGHT dataset version released for the CVPR 2026 challenge~\cite{chen2025bright}, a heterogeneous EO-SAR disaster mapping setting with 1024$\times$1024 pre-event optical and post-event SAR image pairs. Unlike LEVIR-CD and WHU-CD, BRIGHT includes building-status labels and pairs different sensing modalities across time, so both models are adapted from binary optical CD to four-class disaster mapping with background, intact, damaged, and destroyed classes. For BRIGHT, both models are trained on the released training split and evaluated on the released validation split; these BRIGHT numbers are validation-split measurements rather than challenge test-server results. For BRIGHT, we use per-class IoU and AP, target-class mean IoU (tc-mIoU), and target-class mean AP (tc-mAP), where the target-class means are computed only over intact, damaged, and destroyed, excluding background.

\subsection{Experimental Setup}
Experiments are conducted on NVIDIA RTX A6000 and RTX 5090 GPUs. 
We train FAF-CD for 150 epochs using AdamW and use settings aligned with NeXt2Former-CD, with small adjustments to accommodate our architecture and improve optimization stability.
For LEVIR-CD and WHU-CD, we use the reported NeXt2Former-CD results rather than retraining that baseline in this study.
We adopt the public preprocessing and train/val/test splits used by prior work for fair comparison~\cite{paranjape_mamba-based_2024}. 
For LEVIR-CD and WHU-CD, the Siamese encoder is initialized with DINOv3-pretrained ConvNeXt-L weights trained on LVD-1689M~\cite{simeoni_dinov3_2025}.
For BRIGHT, we train a matched NeXt2Former-CD adaptation and FAF-CD using DINOv3-pretrained ConvNeXt-Base instead of ConvNeXt-L to make the comparison feasible under the 1024$\times$1024 EO-SAR setting.
The BRIGHT variants also switch to pseudo-Siamese pre/post encoders for different modalities; the post-event SAR branch is read as a single-channel grayscale input, normalized separately, and replicated to three channels before DINOv3 encoding.
For the BRIGHT task, both models further use a four-class prediction head and class-balanced training losses for the target building-status classes.
During fine-tuning, we use a reduced learning rate for the encoder (0.1$\times$ of the base learning rate); the encoder can be optionally frozen for the first 5 epochs to reduce early-stage instability.

For LEVIR-CD and WHU-CD, we report cF1 and cIoU for the change class, along with overall pixel accuracy (OA), defined as the fraction of correctly classified pixels over the test set.
These binary CD metrics are computed on the test split following common practice in recent remote sensing CD benchmarks~\cite{paranjape_mamba-based_2024, wang_next2former-cd_2026}, while BRIGHT metrics are computed on the validation split.
On the LEVIR-CD~\cite{chen_spatial-temporal_2020} dataset, FAF-CD training with batch size 8 takes 13.47 hours using a single RTX 5090 GPU; for reference, the reported NeXt2Former-CD runtime is 15.44 hours.

\subsection{Results}

\paragraph{Quantitative Evaluation.}
We compare FAF-CD with representative baselines in \cref{tab:mcd_sota_comparison}.
On LEVIR-CD~\cite{chen_spatial-temporal_2020}, FAF-CD achieves the best overall performance with 0.924 cF1 and 0.859 cIoU, improving over NeXt2Former-CD~\cite{wang_next2former-cd_2026} by +0.003 cF1 and +0.005 cIoU.
Although the margin is modest, the improvement is consistent across cF1/cIoU/OA, indicating that the proposed fusion reduces hard false alarms without sacrificing overall pixel correctness.
This trend is also consistent with the fusion ablation in \cref{tab:ablation_fusion}, where naive concatenation yields marginal gains while tri-branch gated aggregation performs best.
On WHU-CD~\cite{ji_fully_2019}, FAF-CD attains 0.955 cF1 and 0.914 cIoU, matching the top-performing hybrid baseline with comparable OA (99.642\% vs.\ 99.646\%).
On BRIGHT validation~\cite{chen2025bright}, FAF-CD also improves tc-mIoU/tc-mAP and OA over NeXt2Former-CD.
The clean results establish competitive accuracy, while the perturbation study in \cref{tab:pseudo_change_robustness} further evaluates robustness on the binary optical CD benchmarks and BRIGHT validation stress tests.

\begin{table*}[t]
  \centering
  \small
  \renewcommand{\arraystretch}{1.08}
  \setlength{\tabcolsep}{3pt}
  \resizebox{\textwidth}{!}{%
  \begin{tabular}{l cc cc cc}
    \toprule
    & \multicolumn{2}{c}{LEVIR-CD test~\cite{chen_spatial-temporal_2020}}
    & \multicolumn{2}{c}{WHU-CD test~\cite{ji_fully_2019}}
    & \multicolumn{2}{c}{BRIGHT val~\cite{chen2025bright}} \\
    \cmidrule(lr){2-3}\cmidrule(lr){4-5}\cmidrule(lr){6-7}
    Method
    & Clean cIoU & Pert. cIoU/cF1
    & Clean cIoU & Pert. cIoU/cF1
    & Clean tc-mIoU/tc-mAP & Pert. tc-mIoU/tc-mAP \\
    \midrule
    M-CD~\cite{paranjape_mamba-based_2024}
    & 85.02 & 84.38/91.53
    & 91.12 & 90.66/95.10
    & -- & -- \\
    NeXt2Former-CD~\cite{wang_next2former-cd_2026}
    & 85.37 & 85.06/91.92
    & \textbf{91.40} & 91.05/95.31
    & 54.74/74.79 & 54.09/73.99 \\
    \textbf{Ours}
    & \textbf{85.88} & \textbf{85.52/92.19}
    & 91.38 & \textbf{91.15/95.37}
    & \textbf{58.46/76.91} & \textbf{58.00/76.52} \\
    \bottomrule
  \end{tabular}}
  \caption{
    Robustness under perturbation stress tests.
    Clean and Pert. report clean and perturbed performance, respectively; binary CD entries use test-set cIoU/cF1, while BRIGHT uses validation-set tc-mIoU/tc-mAP over intact, damaged, and destroyed.
    FAF-CD improves the perturbed averages across both binary optical CD benchmarks and the matched BRIGHT validation stress tests.
    All values are percentages; detailed averaging protocols are provided in the supplement.
  }
  \label{tab:pseudo_change_robustness}
\end{table*}

\paragraph{Perturbation Robustness.}
To directly evaluate robustness to nuisance appearance variations, we conduct a pseudo-change perturbation study on LEVIR-CD and WHU-CD in which only one temporal image is perturbed while the other remains clean.
This setting simulates illumination, color, haze, and shadow shifts that can be mistaken for structural changes in binary optical CD.
As shown in \cref{tab:pseudo_change_robustness}, FAF-CD achieves the highest average perturbed cIoU/cF1 on both LEVIR-CD and WHU-CD among M-CD and NeXt2Former-CD, and the highest perturbed tc-mIoU/tc-mAP among the matched BRIGHT validation variants evaluated.
Because BRIGHT pairs can use different pre/post modalities, its results are reported as a separate modality-side stress-test average rather than being mixed into the LEVIR-CD/WHU-CD pseudo-change protocol.
Together, these results support that frequency-aware fusion improves robustness to nuisance degradations while preserving clean accuracy.
Detailed protocol settings and per-perturbation breakdowns are provided in \cref{sec:supp_pseudo_change_robustness,sec:supp_bright_val_results}.

\begin{figure*}
  \centering
\includegraphics[width=0.524\linewidth]{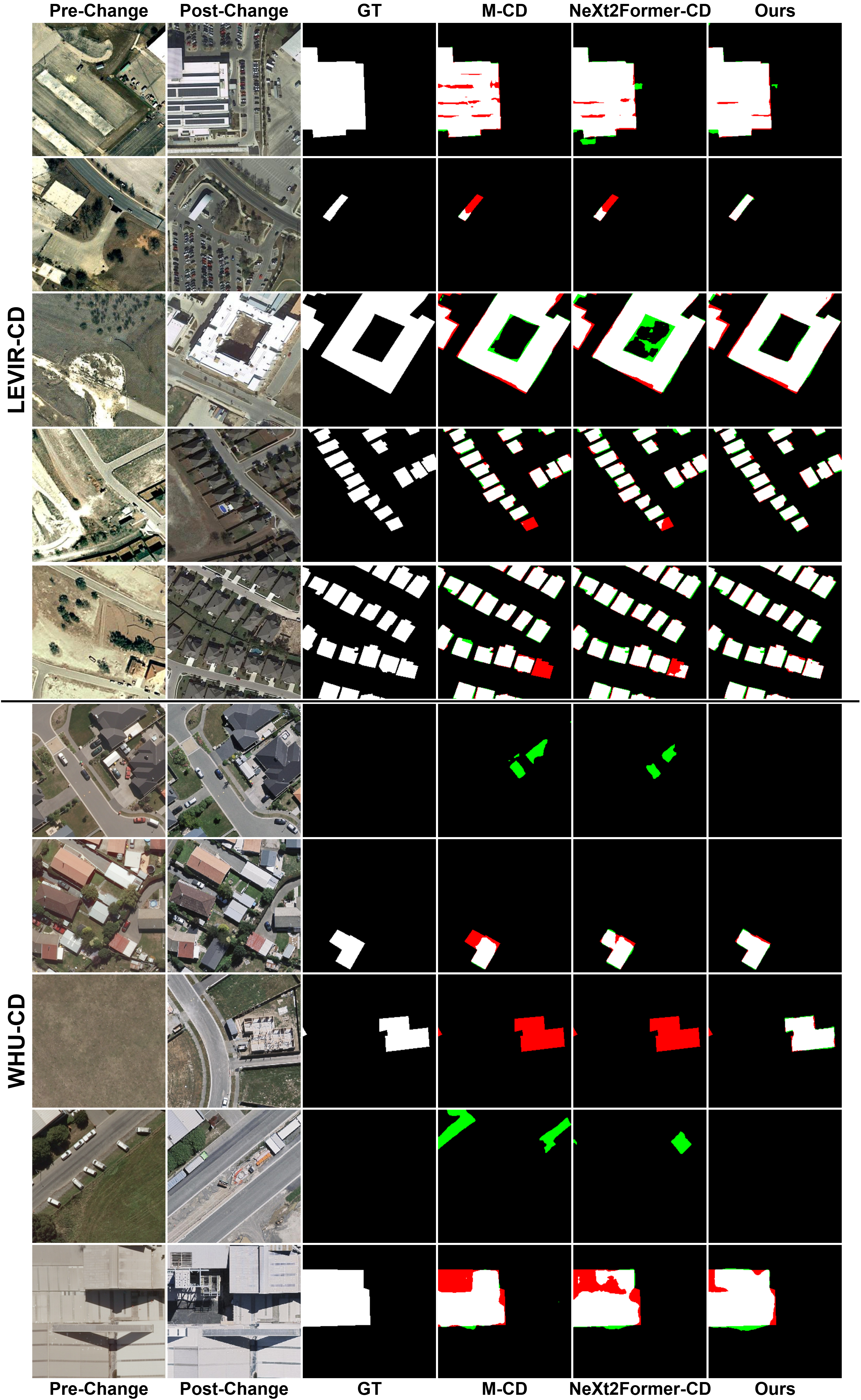}
  \caption{Qualitative results on the LEVIR-CD~\cite{chen_spatial-temporal_2020} and WHU-CD~\cite{ji_fully_2019} datasets. White represents true positives, black represents true negatives, green represents false positives and red represents false negatives.}
  \label{fig:qualitative}
\end{figure*}

\paragraph{Qualitative Analysis.}
Visual results in \cref{fig:qualitative} further support the quantitative improvements. On LEVIR-CD, FAF-CD reduces scattered false positives around shadows, reflective rooftops, and textured backgrounds while producing more coherent building footprints. On WHU-CD, it yields cleaner unchanged regions and fewer fragmented boundaries in cluttered urban scenes. The BRIGHT qualitative comparison, shown in the supplement (\cref{fig:bright_qualitative}), similarly indicates better suppression of over-extended building-status predictions and fewer intact/damaged/destroyed class confusions in multimodal pre/post imagery. Overall, the multi-domain fusion improves boundary fidelity and suppresses pseudo-changes under challenging appearance variations.

\begin{table}[t]
  \centering
  \renewcommand{\arraystretch}{1.15}
  \setlength{\tabcolsep}{4pt}
  \begin{tabular}{p{0.53\columnwidth} c c c}
    \toprule
    Fusion Strategy 
    & cF1 $\uparrow$
    & cIoU $\uparrow$
    & OA $\uparrow$ \\
    \midrule
    FFM
      & 0.923
      & 0.8566
      & 99.358 \\
    FFT
      & 0.922
      & 0.8556
      & 99.351 \\
    DWT
      & 0.921
      & 0.8529
      & 99.336 \\
    \midrule
    FFM + FFT, concat, 1x1 proj
      & 0.923
      & 0.8570
      & 99.357 \\
    FFM + DWT, concat, 1x1 proj
      & 0.923
      & 0.8578
      & 99.361 \\
    \midrule
    \textbf{Tri-Branch Fusion (ours)}
      & \textbf{0.925} 
      & \textbf{0.8606} 
      & \textbf{99.375} \\
    \bottomrule
  \end{tabular}
    \caption{Ablation study on feature fusion strategy on LEVIR-CD~\cite{chen_spatial-temporal_2020} with batch size 8, evaluated on the validation split.}
  \label{tab:ablation_fusion}
\end{table}

\paragraph{Ablation Study.}
To understand the individual contributions within our fusion strategy, we conducted ablations on the LEVIR-CD~\cite{chen_spatial-temporal_2020} validation set (\cref{tab:ablation_fusion}). The baseline configuration uses only the spatial Feature Fusion Module (FFM) following NeXt2Former-CD~\cite{wang_next2former-cd_2026}, which already provides a strong cF1 score of 0.923. Substituting the FFM entirely with either the global FFT or local Haar DWT branches slightly degrades performance. More importantly, simply concatenating the spatial FFM with a single frequency branch (FFM + FFT or FFM + DWT) fails to yield meaningful gains. However, dynamically gating all three representations together in our proposed Tri-Branch module pushes the cF1 score to 0.925. This indicates that spatial, global Fourier, and local wavelet features capture distinct, mutually beneficial information that requires an adaptive aggregation mechanism to be fully exploited.

\paragraph{Computational Efficiency.}
\cref{tab:efficiency_comparison} reports the computational cost of our design. Adding multiple fusion branches increases the parameter count to 417.84M. However, the fvcore-counted GFLOPs are lower than the purely spatial NeXt2Former-CD~\cite{wang_next2former-cd_2026} (135.88 GFLOPs vs. 159.54 GFLOPs). The inference time on a single RTX 5090 is 34.01 ms per image pair, which is faster than NeXt2Former-CD~\cite{wang_next2former-cd_2026} but slower than the lighter M-CD~\cite{paranjape_mamba-based_2024} architecture. These results show that the added frequency branches do not increase the measured inference cost in this setting.

Because BRIGHT uses a different 1024$\times$1024 multimodal adaptation, we profile the BRIGHT variants separately in \cref{tab:bright_efficiency_comparison}. Under the matched ConvNeXt-Base setting, FAF-CD has fewer trainable parameters, lower fvcore-counted GFLOPs, lower inference latency, and lower CUDA peak allocated memory than the adapted NeXt2Former-CD baseline. This high-resolution advantage is consistent with the decoder design: FAF-CD uses a VMamba-based decoder, whereas NeXt2Former-CD retains a Mask2Former decoder, making the efficiency gap more important as the image size increases.

\begin{table}[t]
\centering
\small
\renewcommand{\arraystretch}{1.05}
\setlength{\tabcolsep}{3pt}
\resizebox{\columnwidth}{!}{%
\begin{tabular}{l c c c}
\toprule
Method 
& Trainable Params (M)
& GFLOPs
& Time (ms) \\
\midrule
M-CD  & 69.81 & 28.23 & 33.84 \\
NeXt2Former-CD  & 392.00 & 159.54 & 36.79 \\
Ours  & 417.84 & 135.88 & 34.01 \\
\bottomrule
\end{tabular}}
\caption{
Efficiency comparison between our proposed method, M-CD~\cite{paranjape_mamba-based_2024}, and NeXt2Former-CD~\cite{wang_next2former-cd_2026}. 
Params, GFLOPs, and inference time are measured for one $256\times256$ image pair, matching the LEVIR-CD/WHU-CD patch setting.
The NeXt2Former-CD and FAF-CD entries use the DINOv3 ConvNeXt-L backbone; this table does not use the BRIGHT ConvNeXt-Base adaptation.
Inference time is measured on an RTX 5090 GPU.
}
\label{tab:efficiency_comparison}
\end{table}

\begin{table}[t]
\centering
\small
\renewcommand{\arraystretch}{1.05}
\setlength{\tabcolsep}{3pt}
\resizebox{\columnwidth}{!}{%
\begin{tabular}{l c c c c}
\toprule
Method
& Trainable Params (M)
& GFLOPs
& Time (ms)
& Peak Alloc. (GB) \\
\midrule
NeXt2Former-CD  & 312.79 & 1599.80 & 123.72 & 3.17 \\
FAF-CD  & \textbf{273.88} & \textbf{973.67} & \textbf{108.82} & \textbf{2.48} \\
\bottomrule
\end{tabular}}
\caption{
BRIGHT efficiency comparison between the adapted NeXt2Former-CD baseline and FAF-CD.
All values are measured for one $1024\times1024$ BRIGHT-size image pair with batch size 1 and FP32 inference on an RTX 5090 GPU.
Both entries use the BRIGHT pseudo-Siamese DINOv3 ConvNeXt-Base adaptation.
Time is averaged over 20 timed forward passes after 5 warmup passes; Peak Alloc. reports CUDA peak allocated memory.
GFLOPs are counted by fvcore and should be interpreted comparatively because custom CUDA operators have partial FLOP-counter support.
}
\label{tab:bright_efficiency_comparison}
\end{table}

\section{Conclusion}
\label{sec:conclusion}

We presented \textbf{FAF-CD}, a frequency-aware hybrid framework for remote sensing change detection. FAF-CD combines a DINOv3-initialized ConvNeXt encoder with a rectification-aware tri-branch fusion module that integrates deformable spatial alignment, Fourier-based global discrepancy modeling, and Haar wavelet-based boundary cues. An adaptive gating mechanism aggregates these complementary representations across scales, and a VMamba-based decoder enables efficient global reasoning for dense mask prediction. FAF-CD achieves improved performance on LEVIR-CD, competitive clean accuracy on WHU-CD, and stronger robustness under pseudo-change-aligned perturbations on both binary optical CD datasets. On BRIGHT validation, the matched multimodal FAF-CD adaptation also improves clean and perturbed tc-mIoU/tc-mAP over the adapted NeXt2Former-CD baseline under modality-side stress tests. Ablation studies further show that the gains primarily come from jointly exploiting spatial, Fourier, and wavelet cues with adaptive aggregation rather than single-domain fusion or naive concatenation. Future work includes extending the frequency-aware fusion to additional multimodal datasets and exploring lightweight, learnable frequency operators for further improving robustness under severe appearance and modality shifts.

\section*{Acknowledgments}
This material is based upon work supported by the National Science Foundation under Award No.~2401835. Any opinions, findings, and conclusions or recommendations expressed in this material are those of the author(s) and do not necessarily reflect the views of the National Science Foundation.

{
    \small
    \bibliographystyle{ieeenat_fullname}
    % \bibliography{main}
    \bibliography{CVPR2026_ChangeDetection}
}

% Supplementary material
\clearpage
\maketitlesupplementary

\section{Pseudo-change Robustness Study}
\label{sec:supp_pseudo_change_robustness}

\subsection{Protocol}
\label{sec:supp_robustness_protocol}

We test robustness to pseudo-changes by perturbing one image in each bi-temporal test pair and leaving the other image unchanged. This isolates asymmetric acquisition effects, such as illumination, color balance, haze, and cast shadows, that can resemble structural change.

On LEVIR-CD~\cite{chen_spatial-temporal_2020} and WHU-CD~\cite{ji_fully_2019}, we evaluate M-CD~\cite{paranjape_mamba-based_2024}, NeXt2Former-CD~\cite{wang_next2former-cd_2026}, and FAF-CD under brightness/contrast, color jitter, haze, and shadow perturbations. NeXt2Former-CD is evaluated using its reported LEVIR-CD/WHU-CD configuration, without additional retraining in this study. Each perturbation is applied at severity levels 1--5, separately to the first or second temporal image. The cIoU and cF1 values reported in the main paper average over these one-sided perturbed settings only; clean pairs, both-image perturbations, and noise/blur degradations are not included. All runs use the same test split and the same corrected legacy-compatible evaluation as the clean benchmark.

\begin{figure*}[t]
  \centering
  \includegraphics[width=\textwidth]{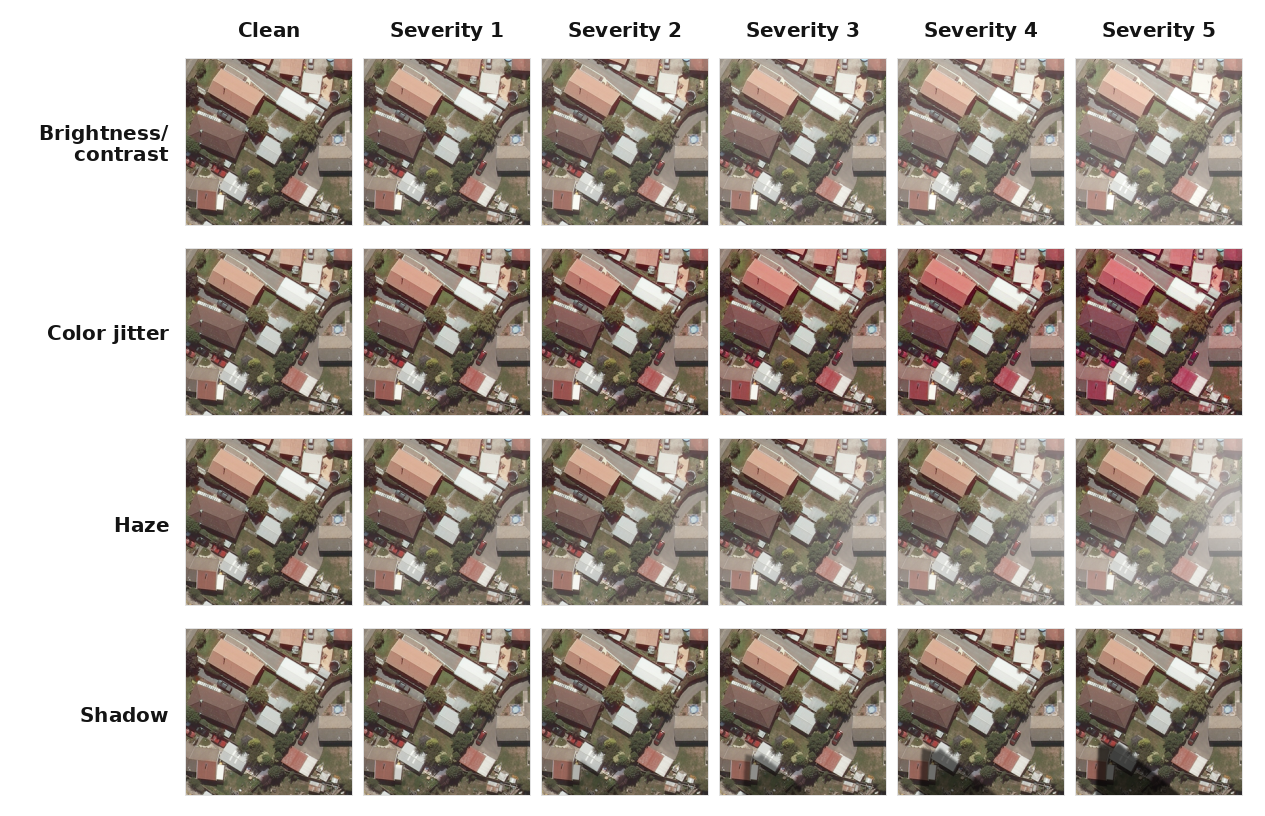}
  \caption{Severity sweep for the pseudo-change perturbations. The same clean WHU-CD~\cite{ji_fully_2019} patch is shown under brightness/contrast, color jitter, haze, and shadow perturbations. In evaluation, each perturbation is applied to only one temporal image at a time.}
  \label{fig:supp_perturbation_severity_sweep}
\end{figure*}

\subsection{Per-perturbation Breakdown}
\label{sec:supp_robustness_breakdown}

\cref{tab:supp_robustness_levir_breakdown,tab:supp_robustness_whu_breakdown} give the per-perturbation results used to compute the aggregate robustness table in the main paper. Each entry is cIoU/cF1 averaged over severity levels 1--5 for the indicated perturbation and temporal side.

\begin{table*}[t]
  \centering
  \small
  \renewcommand{\arraystretch}{1.08}
  \setlength{\tabcolsep}{5pt}
  \begin{tabular}{l c c c c}
    \toprule
    Perturbation & Side & M-CD & NeXt2Former-CD & FAF-CD \\
    \midrule
    brightness/contrast & A & 84.91/91.84 & 85.30/92.07 & 85.79/92.35 \\
    brightness/contrast & B & 83.49/90.99 & 84.47/91.58 & 84.73/91.73 \\
    color jitter & A & 84.92/91.84 & 85.32/92.08 & 85.81/92.36 \\
    color jitter & B & 84.65/91.69 & 85.28/92.05 & 85.74/92.32 \\
    haze & A & 84.89/91.83 & 85.28/92.06 & 85.77/92.34 \\
    haze & B & 83.10/90.76 & 84.65/91.69 & 85.12/91.96 \\
    shadow & A & 84.96/91.87 & 85.19/92.00 & 85.78/92.35 \\
    shadow & B & 84.16/91.39 & 84.96/91.87 & 85.38/92.11 \\
    \bottomrule
  \end{tabular}
  \caption{Per-perturbation robustness results on LEVIR-CD~\cite{chen_spatial-temporal_2020}. Values are cIoU/cF1 averaged over severity levels 1--5.}
  \label{tab:supp_robustness_levir_breakdown}
\end{table*}

\begin{table*}[t]
  \centering
  \small
  \renewcommand{\arraystretch}{1.08}
  \setlength{\tabcolsep}{5pt}
  \begin{tabular}{l c c c c}
    \toprule
    Perturbation & Side & M-CD & NeXt2Former-CD & FAF-CD \\
    \midrule
    brightness/contrast & A & 90.80/95.18 & 91.17/95.38 & 91.16/95.37 \\
    brightness/contrast & B & 89.54/94.47 & 89.71/94.57 & 90.47/94.99 \\
    color jitter & A & 91.16/95.37 & 91.24/95.42 & 91.24/95.42 \\
    color jitter & B & 90.82/95.19 & 91.23/95.41 & 91.37/95.49 \\
    haze & A & 91.03/95.30 & 91.39/95.50 & 91.13/95.36 \\
    haze & B & 90.44/94.98 & 91.14/95.36 & 91.33/95.47 \\
    shadow & A & 91.05/95.32 & 91.36/95.48 & 91.37/95.49 \\
    shadow & B & 90.41/94.96 & 91.17/95.38 & 91.13/95.36 \\
    \bottomrule
  \end{tabular}
  \caption{Per-perturbation robustness results on WHU-CD~\cite{ji_fully_2019}. Values are cIoU/cF1 averaged over severity levels 1--5.}
  \label{tab:supp_robustness_whu_breakdown}
\end{table*}

\subsection{BRIGHT Modality-Side Stress Tests}
\label{sec:supp_bright_val_results}

We additionally evaluate NeXt2Former-CD~\cite{wang_next2former-cd_2026} and FAF-CD on the BRIGHT dataset version released for the CVPR 2026 challenge~\cite{chen2025bright}. BRIGHT uses 1024$\times$1024 pre-event optical and post-event SAR image pairs, so we train matched adaptations of both models for this heterogeneous EO-SAR disaster mapping setting. Both models are trained on the released training split and evaluated on the released validation split; the reported BRIGHT results are therefore validation-split measurements rather than challenge test-server results. Concretely, the BRIGHT variants use pseudo-Siamese pre/post encoders, a 1024$\times$1024 evaluation crop, four output classes (background, intact, damaged, destroyed), and class-balanced training losses for the target building-status classes. The post-event SAR image is handled as a single-channel grayscale modality with its own normalization and is replicated to three channels before DINOv3 encoding. To keep the comparison matched under this higher-resolution setting, both NeXt2Former-CD and FAF-CD use a DINOv3-pretrained ConvNeXt-Base encoder; NeXt2Former-CD keeps the Mask2Former decoder, while FAF-CD uses the VMamba decoder with gated spatial/Fourier/wavelet fusion. Object-detection-style AP/mAP follow the BRIGHT evaluation over the target classes.

\cref{tab:supp_bright_clean_classwise} gives the clean class-wise BRIGHT validation results. Because BRIGHT is heterogeneous EO-SAR, we do not treat the BRIGHT perturbation rows as the same pseudo-change robustness protocol used for LEVIR-CD and WHU-CD, and they are not BRIGHT test-split results or a cross-modal perturbation benchmark. Instead, \cref{tab:supp_bright_perturbation_breakdown} reports controlled modality-side stress tests on the BRIGHT validation split with 697 samples. For the pre-event optical side (Side A), we apply brightness/contrast, color jitter, haze, and shadow perturbations. For the post-event SAR side (Side B), we apply speckle noise, radiometric shift, and resolution blur. All perturbations use severity levels 1--5, giving 20 Side-A rows and 15 Side-B rows per model. We omit optical color perturbations on Side B because the SAR branch is single-channel, so color jitter would have no effect and only act as a sanity check.

\begin{figure*}[t]
  \centering
  \includegraphics[width=0.82\linewidth]{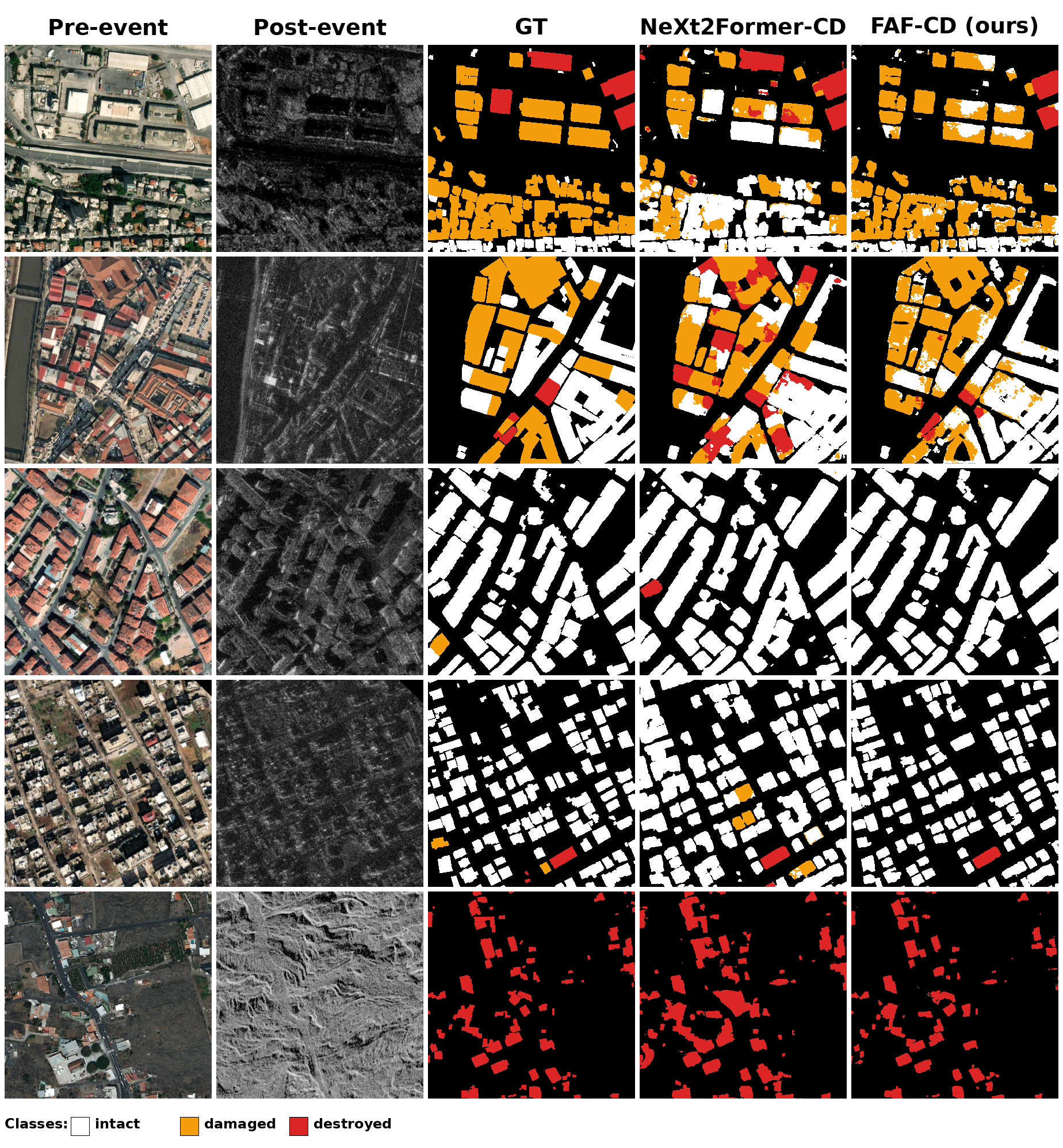}
  \caption{Qualitative BRIGHT validation comparison~\cite{chen2025bright} between the matched NeXt2Former-CD adaptation and FAF-CD. White, orange, and red denote intact, damaged, and destroyed buildings, respectively; black denotes background.}
  \label{fig:bright_qualitative}
\end{figure*}

\begin{table*}[t]
  \centering
  \small
  \renewcommand{\arraystretch}{1.08}
  \setlength{\tabcolsep}{5pt}
  \begin{tabular}{l c c c c}
    \toprule
    Method & Intact IoU/AP & Damaged IoU/AP & Destroyed IoU/AP & tc-mIoU/tc-mAP \\
    \midrule
    NeXt2Former-CD~\cite{wang_next2former-cd_2026}
    & 74.19/93.84
    & 32.47/48.42
    & 57.57/82.10
    & 54.74/74.79 \\
    FAF-CD
    & \textbf{80.05/94.83}
    & \textbf{35.24/52.49}
    & \textbf{60.10/83.39}
    & \textbf{58.46/76.91} \\
    \bottomrule
  \end{tabular}
  \caption{
    Clean BRIGHT validation class-wise results~\cite{chen2025bright}.
    tc-mIoU/tc-mAP is computed over the BRIGHT target classes intact, damaged, and destroyed, excluding background.
    All values are percentages.
  }
  \label{tab:supp_bright_clean_classwise}
\end{table*}

\begin{table*}[t]
  \centering
  \scriptsize
  \renewcommand{\arraystretch}{1.05}
  \setlength{\tabcolsep}{2pt}
  \resizebox{\textwidth}{!}{%
  \begin{tabular}{l c l c c c c c c}
    \toprule
    Perturbation & Side & Method & S1 & S2 & S3 & S4 & S5 & Avg. \\
    \midrule
    \multicolumn{9}{l}{\emph{Side A: pre-event optical branch}} \\
    \midrule
    brightness/contrast & A & NeXt2Former-CD & 54.75/74.95/69.10 & 54.82/74.90/69.16 & 54.79/74.68/69.13 & 54.43/73.95/68.78 & 54.19/73.53/68.66 & 54.59/74.40/68.96 \\
     &  & \textbf{FAF-CD} & \textbf{58.51/77.10/72.09} & \textbf{58.51/77.25/72.11} & \textbf{58.24/77.10/71.87} & \textbf{57.80/76.60/71.51} & \textbf{56.78/75.75/70.68} & \textbf{57.97/76.76/71.65} \\
    \midrule
    color jitter & A & NeXt2Former-CD & 54.72/74.70/69.08 & 54.65/74.64/69.03 & 54.54/74.55/68.93 & 54.43/74.61/68.83 & 54.14/74.28/68.54 & 54.49/74.56/68.88 \\
     &  & \textbf{FAF-CD} & \textbf{58.43/76.83/72.01} & \textbf{58.45/76.80/72.04} & \textbf{58.35/76.66/71.96} & \textbf{58.38/76.63/72.02} & \textbf{58.14/76.27/71.82} & \textbf{58.35/76.64/71.97} \\
    \midrule
    haze & A & NeXt2Former-CD & 54.57/74.72/68.95 & 54.38/74.64/68.79 & 54.13/74.42/68.56 & 53.85/74.20/68.33 & 53.39/73.76/67.92 & 54.07/74.35/68.51 \\
     &  & \textbf{FAF-CD} & \textbf{58.72/77.02/72.31} & \textbf{58.83/77.09/72.43} & \textbf{58.79/77.13/72.42} & \textbf{58.66/77.07/72.32} & \textbf{58.48/76.96/72.17} & \textbf{58.70/77.05/72.33} \\
    \midrule
    shadow & A & NeXt2Former-CD & 54.70/74.75/69.05 & 54.69/74.70/69.04 & 54.71/74.72/69.06 & 54.71/74.57/69.06 & 54.69/74.45/69.07 & 54.70/74.64/69.06 \\
     &  & \textbf{FAF-CD} & \textbf{58.43/76.89/72.01} & \textbf{58.41/76.85/71.99} & \textbf{58.33/76.86/71.94} & \textbf{58.24/76.89/71.85} & \textbf{57.99/76.80/71.66} & \textbf{58.28/76.86/71.89} \\
    \midrule
    \multicolumn{9}{l}{\emph{Side B: post-event SAR branch}} \\
    \midrule
    speckle noise & B & NeXt2Former-CD & 54.70/74.44/69.07 & 54.67/74.18/69.03 & 54.09/73.55/68.57 & 51.49/71.82/66.33 & 47.41/67.22/62.63 & 52.47/72.24/67.13 \\
     &  & \textbf{FAF-CD} & \textbf{58.15/76.62/71.78} & \textbf{57.99/76.46/71.65} & \textbf{57.61/76.11/71.33} & \textbf{56.78/75.42/70.60} & \textbf{56.01/74.70/69.92} & \textbf{57.31/75.86/71.06} \\
    \midrule
    radiometric shift & B & NeXt2Former-CD & 54.70/74.90/69.05 & 54.64/74.86/68.99 & 54.49/74.78/68.87 & 53.61/74.30/68.14 & 52.11/72.61/66.86 & 53.91/74.29/68.38 \\
     &  & \textbf{FAF-CD} & \textbf{58.45/76.92/72.04} & \textbf{58.45/76.97/72.06} & \textbf{58.40/76.99/72.02} & \textbf{58.42/77.10/72.06} & \textbf{57.94/76.65/71.71} & \textbf{58.33/76.93/71.98} \\
    \midrule
    resolution blur & B & NeXt2Former-CD & 54.74/73.99/69.03 & 54.81/73.80/69.10 & 54.46/73.50/68.75 & 54.01/73.09/68.35 & 53.81/72.85/68.21 & 54.36/73.44/68.69 \\
     &  & \textbf{FAF-CD} & \textbf{57.65/76.19/71.29} & \textbf{57.26/75.96/70.92} & \textbf{57.11/75.48/70.82} & \textbf{57.06/75.42/70.90} & \textbf{56.08/74.71/70.03} & \textbf{57.03/75.55/70.79} \\
    \bottomrule
  \end{tabular}}
  \caption{
    Severity-wise BRIGHT validation modality-side stress-test results~\cite{chen2025bright}.
    Each cell is tc-mIoU/tc-mAP/tc-mF1 over the BRIGHT target classes intact, damaged, and destroyed, excluding background.
    S1--S5 denote perturbation severity levels, and Avg. is averaged over severity levels 1--5 for the indicated side.
    Side A rows degrade the pre-event optical branch, while Side B rows degrade the post-event SAR branch.
    These rows are not BRIGHT test-split results; the main paper reports their compact average as a separate BRIGHT validation stress-test column rather than mixing them into the LEVIR-CD/WHU-CD pseudo-change protocol.
    Bold values indicate the better result between methods under the same perturbation setting.
    All values are percentages.
  }
  \label{tab:supp_bright_perturbation_breakdown}
\end{table*}

\begin{figure*}[t]
  \centering
  \begin{subfigure}[t]{0.48\textwidth}
    \centering
    \includegraphics[width=0.88\linewidth]{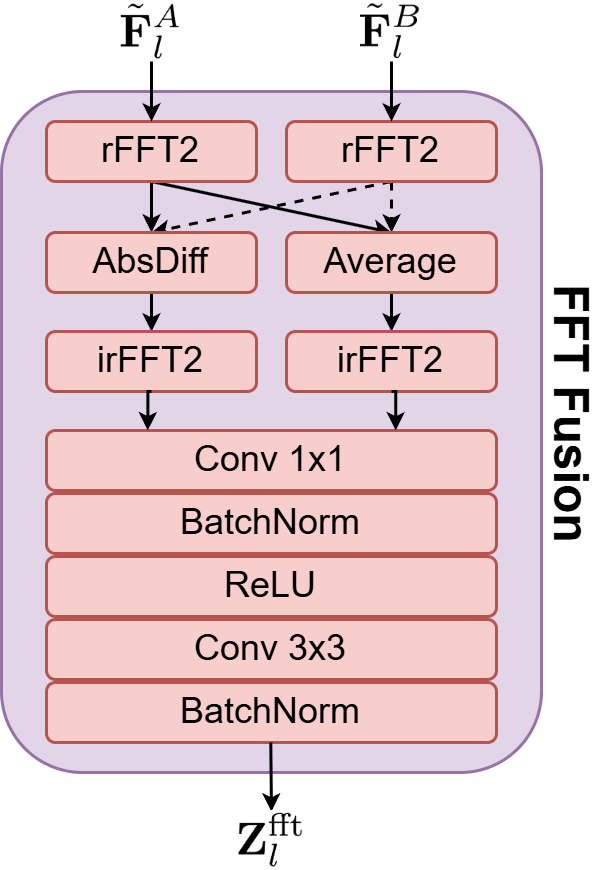}
    \caption{FFT Fusion branch. The input features are transformed with rFFT2, compared through spectral average and absolute difference, and mapped back with irFFT2 before convolutional refinement.}
    \label{fig:fft_fusion}
  \end{subfigure}
  \hfill
  \begin{subfigure}[t]{0.48\textwidth}
    \centering
    \includegraphics[width=0.88\linewidth]{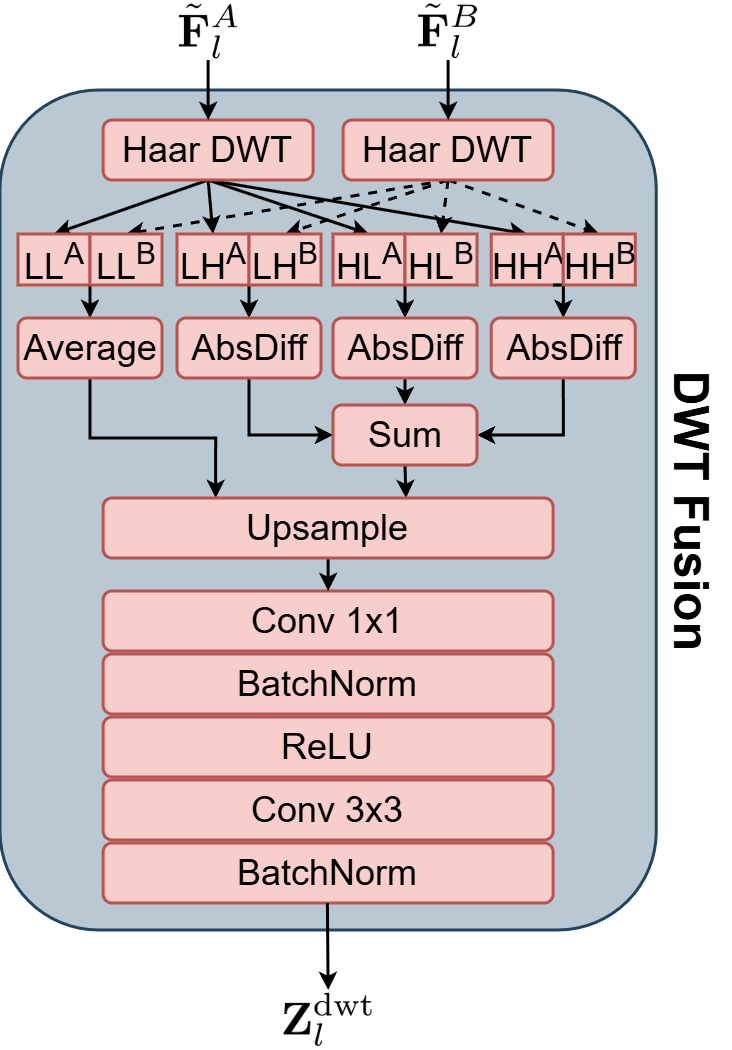}
    \caption{DWT Fusion branch. Haar 2D-DWT decomposes features into low- and high-frequency subbands; high-frequency differences emphasize structural boundaries while low-frequency averages preserve shared context.}
    \label{fig:dwt_fusion}
  \end{subfigure}
  \caption{Internal structures of the frequency-domain fusion branches.}
  \label{fig:supp_frequency_branch_details}
\end{figure*}

\end{document}